\documentclass[]{article}
\usepackage{proceed2e}

\title{Multi-Task Regularization with Covariance Dictionary for \\Linear Classifiers}
\usepackage{graphicx}

\author{} 

\author{ {\bf Fanyi Xiao} \\
School of Computer Science\\
Carnegie Mellon University\\
Pittsburgh, PA 15213 \\
\And
{\bf Ruikun Luo}  \\
Department of Mechanical Engineering\\
Carnegie Mellon University\\
Pittsburgh, PA 15213 \\
\And
{\bf Zhiding Yu}   \\
Department of Electrical \\ and Computer Engineering\\
Carnegie Mellon University\\
Pittsburgh, PA 15213 \\
}

\begin{document}

\maketitle

\begin{abstract}
In this paper we propose a multi-task linear classifier learning problem called D-SVM (Dictionary SVM). D-SVM uses a \emph{dictionary} of parameter covariance shared by all tasks to do multi-task knowledge transfer among different tasks. We formally define the learning problem of D-SVM and show two interpretations of this problem, from both the probabilistic and kernel perspectives. From the probabilistic perspective, we show that our learning formulation is actually a MAP estimation on all optimization variables. We also show its equivalence to a multiple kernel learning problem in which one is trying to find a re-weighting kernel for features from a dictionary of basis (despite the fact that only linear classifiers are learned). Finally, we describe an alternative optimization scheme to minimize the objective function and present empirical studies to valid our algorithm.
\end{abstract}
\section{INTRODUCTION}
Recent years have seen rapid development in Multi-Task Learning (MTL) which demonstrates the effectiveness of exploiting the correlation among different tasks, if they are correlated. There have been considerable applications of MTL on different problem domains including computer vision \cite{hwang2011sharing,quattoni2008transfer,salakhutdinov2011learning}, natural language processing \cite{ando2005framework} and anti-spam filters \cite{attenberg2009collaborative}. In previous literatures, there are two main approaches of knowledge transfer between various tasks. The first family of approach is to directly regularize the parameters learned by different tasks, e.g. regularizing the deviation of different task parameters from the ``mean" parameter \cite{evgeniou2004regularized,parameswaran2010large} or enforcing priors that are shared by all tasks \cite{daume2009bayesian}. The second family of popular MTL formulations assume that data from all tasks have similar feature representations which lie in a lower dimensional manifold that can then be discovered by optimizing over certain convex forms\cite{ando2005framework,argyriou2008convex}. More specifically, \cite{argyriou2008convex} assumes that data from different tasks share a feature representation which is lower dimensional than that of the original feature space, thus making it feasible to discard nominal features jointly among different tasks. This can be achieved by adding a regularization term on the $l_{2,1}$ norm of the parameter matrix leading to the group sparsity property. Recently, \cite{parikh2011treegl} proposes to exploiting the graph as a tool to enforce task-relatedness regularization on the graph. \\\\
Unlike previous methods, we propose to learn a dictionary and use that dictionary as a basis for modeling covariances of multi-task parameters. To fit the covariance of our task parameters $w$, sparse coding is used to select entries from the learned dictionary in a sparse manner by solving LASSO which is exactly a least square problem with $l_1$ (sparsity) regularization. We will show that regularizing all task parameters' covariances in this way is actually to assume that the task parameters $w_t,t=1,2,3,...,T$ coming from different Gaussian distributions share the same elements constituting their covariances. Except for the implicit constraints that all tasks share a basis for their covariances, we also try to enforce explicit constraints on the dictionary coefficients $\alpha_t,t=1,2,3,...,T$ to make them as close as possible. \\\\
The dictionary-based multi task learning problem can be written as an optimization problem and we will show how to optimize it using an alternating optimizing scheme. For the covariance selection step, we update the dictionary coefficient for each task as LASSO problems. In the dictionary updating step, we develop a projection gradient descent method to learn the dictionary because of the nonnegative and $l_1$ ball constraints on the dictionary entries. Finally, with known dictionary coefficients for each task, the SVM learning problem becomes a kernel SVM learning problem with a feature re-weighting kernel which can be solved using standard SVM solver like LIBSVM \cite{CC01a}.\\\\
%In many classification problems such as object localization and recognition \cite{felzenszwalb2008discriminatively,malisiewicz-iccv11}, it is crucial to achieve real-time online computation in order to guarantee practical use of the algorithm. On the other hand, people can undertake certain off-line computation time which constitutes a trade-off between training and testing. Linear classifiers are highly effective and efficient, and can be used in real-time scenarios due to its $\mathcal{O}$(n) online computation time which only requires multiplications or convolution operations. Therefore, we only explore the learning problem of linear classification models in this paper.
Our major contribution in this paper lies in the fact of further boosting the performance of linear classifiers by leveraging relatedness among different models by covariance regularization  while maintaining relatively low computation complexities during testing. In many classification problems such as object localization and recognition [11, 18], it is crucial to achieve real-time online computation in order to guarantee practical use of the algorithm. On the other hand, people can undertake certain off-line computation time which constitutes a trade-off between training and testing. Linear classifiers are highly effective and efficient, and can be used in real-time scenarios due to its O(n) online computation time which only requires multiplications or convolution operations. Therefore, we only explore the learning problem of linear classification models in this paper.
\\\\
We demonstrate the effectiveness of our method empirically on some real-world datasets. Firstly, we compare the performance of standard linear SVM and two learning problems we propose in this paper on UCI Arrhythmia dataset and show the superior performance with both of our methods over the standard linear SVM. The results on the MNIST digit recognition for two kinds of experimental setting as class SVMs and instance SVMs are also shown. The results and visualization of our learned parameters demonstrate the intuition for our methods. Finally, we apply our method for image classification tasks on Caltech101 which yields better performance compared with standard linear SVM baseline.\\\\
The paper is organized as follows: In section 2, we give brief overview of related work on multi-task learning and dictionary learning. In section 3, we formally propose two learning problems which exploit multi-task relatedness implicitly and explicitly and interpret them from probabilistic and kernel perspectives. In section 4, we describe how to optimize the learning problems proposed in section 3. Experimental results are given in section 5. We conclude the paper in section 6.
\section{RELATED WORK}
\label{related_work}
The basic idea of Multi-Task Learning (MTL) is based on the assumption that different tasks have intrinsic relatedness which can be exploited to enhance the learning process if we learn them jointly. Below we briefly give two examples of MTL to unveil its intuition. In \cite{evgeniou2004regularized}, multiple tasks are learned in a joint manner by adding regularization term on deviations of parameters of different tasks to the mean parameters
$$\min_{w_0,v_t,\xi_{it}} J(w_0,v_t,\xi_{it})=\sum_{t=1}^T\sum_{i=1}^m\xi_{it}+\frac{\lambda_1}{T}\sum_{t=1}^T||v_t||^2+\lambda_2||w_0||^2$$
$$s.t.~~~y_{it}(w_0+v_t){\cdot}x_{it}\geq1-\xi_{it}$$
$$s.t.~~~\xi_{it}\geq0$$
which forces task parameters to be as close as possible. The formulation adopts the hinge loss function as SVM does. An alternative way of enforcing the relatedness assumption is to select features jointly among different tasks using an $l_{2,1}$ norm which leads to the group sparsity \cite{argyriou2008convex}
$$\varepsilon(A,U)=\sum_{t=1}^T\sum_{i=1}^mL(y_{ti},a_t{\cdot}U^Tx_{ti})+\gamma||A||_{2,1}^2$$
where U is a matrix of feature $u_i$ and A is a feature selection matrix. The above expression yields a solution of A which has many zero elements with some nonzero elements selected by all tasks (this is called the group sparsity property).
\section{MULTI-TASK REGULARIZATION}
\subsection{LARGE-MARGIN LINEAR CLASSIFIER}
\label{large1}
The large-margin based algorithm formulate the learning problem as an maximization problem over the geometric margins in both sides of the classification hyperplane
$$\min_{w}J(w)=\frac{1}{2}w^Tw+C\sum_i{\xi_i}$$
$$s.t.~~~y_i(w^Tx_i+b){\geq}1-\xi_{i}$$
$\xi_i$ are the slack variables which allow data point $x_i$ to penetrate the margin generated by the learning algorithm. In standard learning setting, if we have N tasks to learn in a real application, and we want to learn N large-margin classifiers. We just apply learning algorithms N times to maximize N margins separating our data in N tasks. However, it is highly possible (and in practice often the case) that the N tasks we want to learn are highly correlated with each other and thus we can enforce this as \emph{a prior} into our learning problem and learn all N tasks jointly. \\\\
\subsection{MULTI-TASK COVARIANCE REGULARIZATION FOR LINEAR LARGE-MARGIN CLASSIFIERS}
In this section, we first propose a multi-task learning problem and give an algorithm solving it. Then, we propose an extension to this algorithm. \\\\
Suppose we have T tasks, each task has $N_t$ training data points. We denote the classification parameter for task $t$ as $w_t{\in}$$\mathcal{R}$$^m$. Unlike the two ways to enforce the relatedness between different tasks introduced in Section \ref{related_work}, we model the covariance matrix $\Omega_t$ of $w_t$ and assume that $\Omega_t$ is generated from the same dictionary $B$ and parameterized by $\alpha_t$. Thus, by sharing the same covariance dictionary $B$, $w_t$ can transfer knowledge between different tasks. To simplify the model and to set up a convex optimization problem, it requires that $\Omega_t$ is a positive semi-definite matrix. So there is an additional assumption that $$\Omega_t=Diag(B\cdot\alpha_t)$$ where $\forall\alpha_t\geq0$ and all elements of dictionary $B$: $B_{ij}>0,{\forall}i,j$. Hence, the problem in Section \ref{large1} can be modified as $$\min_{w,\alpha,B}J(w,\alpha,B)=\sum_{t=1}^TJ_t(w_t,\alpha_t,B)$$
$$=\sum_{t=1}^T[\lambda_1\sum_{i=1}^{N_t}\xi_{ti}+\lambda_2w_t^T\Omega_t^{-1}w_t+\gamma||\alpha_t||_1]\eqno{(1)}$$
$$s.t.~~~y_{ti}(w_t^Tx_{ti}+b_t){\geq}1-\xi_{ti}$$
$$s.t.~~~\Omega_t=Diag(B\cdot\alpha_t)$$
$$s.t.~~~B_{ij}\geq0,||B_j||_1\leq1,{\forall}i,j$$
where $\xi_{ti}$ is the slack variable corresponding to the $i^{th}$ data points of task $t$.\\\\
%$D$ is the dictionary we use to generate $\Omega(\alpha_t)\in$$\mathcal{R}$$^{m{\times}m}$, of which all elements $d_{ij}$ must be nonnegative and all entries $d_j$ should reside in an $l_1$ ball constraint.\\\\
We termed the above learning problem as D-SVM. The formulation basically draw elements from the dictionary $B$ in a sparse manner (by the $l_1$ norm on the task dictionary coefficient $\alpha_t$) and combine them linearly to form $\Omega_t$. With the dictionary $B$, we couple all tasks together in a joint optimization problem, which makes it feasible to transfer knowledge among tasks. We will give two perspectives of this formulation in the following part from both \emph{probabilistic} and \emph{kernel} perspectives.
\subsubsection{Probabilistic Perspective: MAP Estimation}
In this section, we view the learning problem eq.(1) from a probabilistic perspective. Now we only consider minimize $J_t$.
$$\min_{w_t,\alpha_t,B}J_t(w_t,\alpha_t,B)=\lambda_1\sum_{i=1}^{N_t}\xi_{ti}+\lambda_2w_t^T\Omega_t^{-1}w_t+\gamma||\alpha_t||_1$$
We denote $D_t$ as the training data for the $t^{th}$ task. In a bayesian view, suppose $B$ is given, the optimal parameter $w_t$ could be found by performing a MAP estimation
$$w_t^*=\arg\max_{w_t}P(w_t,\alpha_t,\delta_t|D_t)$$
$$=\arg\max_{w_t}\frac{P(w_t,\alpha_t,\delta_t)P(D_t|w_t,\alpha_t,\delta_t)}{P(D_t)}$$
where $\delta_t$ is the element on the main diagonal of $\Omega_t$. The above equation is equivalent to minimize the negative log posterior
$$w_t^*=\arg\min_{w_t}-{\log}P(w_t,\alpha_t,\delta_t|D_t)$$
$$=\arg\min_{w_t}-[{\log}P(w_t,\alpha_t,\delta_t)+{\log}P(D_t|w_t,\alpha_t,\delta_t)]$$
Given $w_t$, $\alpha_t$ and $\delta_t$ are independent to $D_t$. Hence, we have
$$-{\log}P(D_t|w_t,\alpha_t,\delta_t)=-{\log}P(D_t|w_t)$$
Here we use hinge loss function as the negative log term of the data likelihood given the parameter $w_t$
$$-{\log}P(D_t|w_t)=L(Y_t,X_t;w_t)=\lambda_1\sum_{i=1}^{N_t}\xi_{ti}$$
We assume the conditional distribution $w_t|\delta_t$ follows a Gaussian distribution of zero mean and covariance matrix $\Omega_t$, the zero mean assumption could be satisfied by manipulating our data, therefore we have
$$w_t|\delta_t{\sim}N(\textbf{0},\Omega_t)$$
Then we can write down the negative log term $-{\log}P(w_t|\delta_t)$ as
$$-{\log}P(w_t|\delta_t)=w_t^T{\Omega_t}^{-1}w_t$$
The above calculation of $w_t$ demonstrates that optimizing the objective function we propose in eq. (1) is actually solving for a MAP estimation for $w_t$ if we assume that it follows a Gaussian distribution with zero mean and known covariance $\Omega_t$ which is composed by elements in a dictionary $B$.\\\\
The process of composing the covariance matrix $\Omega_t$ from dictionary $B$ is actually a \emph{hyperprior} on the covariance which we assume $\delta_t$ follows. When $\alpha_t$ is given, $\delta_t$ is determined with probability 1. Thus
$$-{\log}P(\delta_t|\alpha_t)=0$$
Finally, we assume a marginal distribution on $\alpha_t$
$$-{\log}P(\alpha_t)=\gamma||\alpha_t||_1$$
Thus, from a generative view, we can get the probability of $P(w_t,\alpha_t,\delta_t)=P(w_t|\delta_t)P(\delta_t|\alpha_t)P(\alpha_t)$. By adding the above terms, we can get that
$$w_t^*=\arg\max_{w_t}P(w_t,\alpha_t,\delta_t|D_t)=\arg\min_{w_t}\lambda_1\sum_{i=1}^{N_t}\xi_{ti}$$
$$+w_t^T{\Omega_t}^{-1}w_t+\gamma||\alpha_t||_1=\min_{w_t,\alpha_t,B}J_t(w_t,\alpha_t,B)$$
\subsubsection{Kernel Perspective: Selecting Feature Re-weighting Kernel}
Observing the eq. (1), we can find that when $\alpha_t$ are given, which infers that $\Omega_t$ are given
$$\min_{w_t}J_t(w_t)=\lambda_1\sum_{i=1}^{N_t}\xi_{ti}+\lambda_2w_t^T\Omega_t^{-1}w_t$$
The learning problem becomes an SVM learning problem with a kernel for re-weighting different features
$$\min_{w_t}J_t(w_t)=w_t^T(\lambda_2\Omega_t^{-1})w_t+\lambda_1\sum_{i=1}^{N_t}\xi_{ti}$$
Here we denote the matrix $K=\lambda_2\Omega_t^{-1}$ as a diagonal matrix K, then our learning problem becomes an SVM learning problem with a feature re-weighting kernel K. If we use a new variable $\tilde{w_t}=K^{\frac{1}{2}}w_t$, then our problem becomes
$$\min_{w_t}J_t(\tilde{w_t})=\tilde{w_t}^T\tilde{w_t}+\lambda_1\sum_{i=1}^{N_t}\xi_{ti}$$
Which is now a standard SVM problem. The transformation $\tilde{w_t}=K^{\frac{1}{2}}w_t$ is exactly to multiply our task parameter $w_t$ by a diagonal matrix to reweight our features. In section 5, we will empirically demonstrate the effects of re-weighting our features by the learned matrix $K$. Note that even though our learning problem can be cast as an SVM problem with re-weighting kernel, our learned classifier is definitely a \emph{linear} classifier since we only need an inner product operation between parameter $w$ and the testing sample $x$ to determine the label of $x$. \\\\
In multiple kernel learning setting \cite{bach2004multiple}, the kernel matrix for a particular learning problem is chosen as a conic combination of kernel matrices and the learning problem for the coefficients is formed as a QCQP. Here the learning problem eq. (1) is just a multiple kernel learning problem selecting re-weighting kernels from a dictionary $B$.
\subsubsection{Explicit Mean Regularization on Dictionary Coefficients $\alpha$}
Although we have enforced the multi-task regularization implicitly by making all parameters $w_t$ to share a dictionary from which we form the covariance matrix of the gaussian distribution which we assume $w_t$ will follow. However, we still want to see if the implicit knowledge transferring is effective enough for exploiting the correlations among different models. To see that, we propose to add an explicit parameter mean regularization term into our learning problem (1)
$$\min_{w,\alpha,B}J(w,\alpha,B)=\sum_{t=1}^T\min_{w_t,\alpha_t,B}J_t(w_t,\alpha_t,B)\eqno{(2)}$$
$$=\sum_{t=1}^T\{\lambda_1\sum_{i=1}^{N_t}\xi_{ti}+\lambda_2w_t^T\Omega_t^{-1}w_t+\gamma||\alpha_t||_1+\lambda_3||\alpha_t-\bar{\alpha}||^2\}$$
where $\bar{\alpha}$ is just the mean of all coefficients $\alpha_t$. We call this problem MD-SVM (Mean Regularized D-SVM). The term $\lambda_3||\alpha_t-\bar{\alpha}||^2$ is just a mean regularization term which forces the coefficients $\alpha_t$ for different task $t$ to be as close as possible. Note that adding this term will not give us much difficulties in optimizing the learning problem, what we need to do is to make a small modification in optimizing $\alpha_t$.
\label{lambda3}
\section{OPTIMIZATION}
In this section, we describe the algorithm for optimizing the problem we propose in eq. (1). By observing the problem one can find that it is a convex problem which gives us a global optimal from arbitrary initial points. For $w_t$, the minimization problem is just over a quadratic term with linear constraints. As for $\alpha_t$, we know the $l_1$ norm $||\alpha_t||_1$ is a convex term and the term $w_t^T\Omega_t^{-1}w_t$ is convex on $\alpha_t$ if $\Omega_t$ is positive semidefinite \cite{boyd2004convex}. Also we know that $\Omega_t=Diag(B\cdot\alpha_t)$ which is linear on $\alpha_t$, thus leads to the observation that $J$ is convex on $\alpha_t$. Having the convexity property, we propose an alternative minimization algorithm which is local-minimum free for any starting point.
\subsection{OPTIMIZING HYPERPLANE $w$: STANDARD SVM}
Firstly, we solve the problem of minimizing $J$ over $w$ given $\alpha$ and $B$. When $\alpha$ and $B$ are given, the optimization on $w$ is like the following
$$\min_{w}J_w(w)=\lambda_1\sum_{t=1}^T\sum_{i=1}^{N_t}\xi_{ti}+\lambda_2\sum_{t=1}^Tw_t^T\Omega_t^{-1}w_t$$
$$s.t.~~~y_{ti}(w_t^Tx_{ti}+b_t){\geq}1-\xi_{ti}$$
Denote $K=\lambda_2\Omega_t^{-1}$, note that we can transform our data to make use of the off-the-shelf SVM solver like LIBSVM \cite{CC01a} by $\tilde{X}=K^{-\frac{1}{2}}X$. Also we denote $\tilde{w}=K^{\frac{1}{2}}w$, then the above problem becomes
$$\min_{\tilde{w}}J_w(\tilde{w})=\sum_{t=1}^T||\tilde{w_t}||^2+\lambda_1\sum_{t=1}^T\sum_{i=1}^{N_t}\xi_{ti}$$
$$s.t.~~~y_{ti}(\tilde{w_t}^T\tilde{x_{ti}}+b_t){\geq}1-\xi_{ti}$$
Which can be solved exactly using the standard solver. We can then recover $w_t$ by the equation $w_t=K^{-\frac{1}{2}}\tilde{w_t}$.
\subsection{OPTIMIZING COEFFICIENTS $\alpha$: L1-PENALIZED LEAST SQUARE}
To enforce the constraint $\Omega_t=Diag(B\cdot\alpha_t)$, we transform the constrained optimization problem into an unconstrained problem by a Lagrangian multiplier $\nu$
$$\min_{\alpha_t,\delta_t}J_t(\alpha_t,\delta_t)=\lambda_2w_t^TDiag(\delta_t)^{-1}w_t+\nu[\frac{\gamma}{\nu}||\alpha_t||_1$$
$$+||\delta_t-D\alpha_t||^2]\eqno{(3)}$$
In theory, if we set $\nu$ to be $\infty$, then the unconstrained problem is equivalent to the problem with the equality constraint. In practice, we just set $\nu$ to be a very large value. For the above minimization problem on $\alpha$ and $\delta$, we can also solve it using an alternative method. The above unconstrained formulation gives us the advantage to make use of the off-the-shelf LASSO algorithm like the feature-sign search algorithm \cite{lee2007efficient} to minimize over $\alpha_t$. For $\delta_t$, we can solve it in a closed form by setting $\partial{J}/\partial{\delta_t}=0$. \\\\
Considering the formulation with explicit mean regularization on $\alpha_t$ in eq.(2), we could transform the problem into
$$\min_{\alpha_t}||C\alpha_t-d||^2+\gamma||\alpha_t||_1$$
where
$$Q={\eta}B^TB+\lambda_3(\frac{N-1}{N})^2{\cdot}I$$ $$p={\eta}D^T\delta_t+\lambda_3\frac{N-1}{N^2}(\sum_{j{\neq}i}\alpha_j)$$
with equalities $Q=C^TC$ and $p=C^Td$ which can give us $C$ and $d$ efficiently by an SVD decomposition and then solving a linear system. Thus, we could still use the standard solver for LASSO problem.
\subsection{OPTIMIZING DICTIONARY $B$: GRADIENT DESCENT}
It is very important to have a good dictionary to model the covariance of different tasks' parameters. In this part, we discuss about how to learn the dictionary $B$. From eq. (3), we can observe that to learn $B$, we are actually minimizing a function over $B$
$$\min_BJ_B(B)=\sum_{t=1}^T||\delta_t-B\alpha_t||^2$$
$$s.t.~~~B_{ij}\geq0,||B_j||_1\leq1,{\forall}i,j$$
Here since $B$ is a dictionary of covariance, we constrain elements $B_{ij}$ to be nonnegative. Also, we constrain the $l_1$ norm of dictionary entries $B_j$ to be less or equal than 1, otherwise the solution will be trivial if we have more entries than tasks. To solve this problem, we have derived a projection gradient descend method. In each step, we first calculate the gradient of $J_B$ with regard to $B_j$ as $\nabla_{B_j}$ and then calculate a gradient descend step as
$$\hat{B}_j^{n+1}=B_j^{n}-{\eta}\nabla_{B_j}$$
Where $\eta$ is the step size of the gradient descend. Then, we project $\hat{B}_j^{n+1}$ to a probabilistic simplex  the $l_1$ ball constraint $||B_j||\leq1$ and also the nonnegative constraint $B_{ij}\geq0$. For this projection, there are many efficient algorithms. In this paper, we use the algorithm in \cite{duchi2008efficient} which solves this problem with time complexity.
\section{EMPIRICAL RESULTS}
We present extensive empirical studies to evaluate our algorithm. We test our algorithms, D-SVM and MD-SVM, on three real-world data sets, Arrhythmia dataset, MNIST dataset and Caltech 101 dataset. To evaluate the performance of D-SVM and MD-SVM, we use standard linear SVM and logistic regression as our baseline.
\subsection{ARRHYTHMIA DATASET}
Arrhythmia data set consists of 279 attributes from 452 patients which contains basic information and test signals of the patients. This dataset has 16 classes which indicate 16 arrhythmia types \cite{Frank+Asuncion:2010}.\\\\
We first normalize the data and delete some useless features (with the same value or with almost all missing value). There are 13 classes in the dataset. In addition, there are 5 classes with less than 10 instances which are ignored to avoid overfitting. So after normalization, the dataset consists of 8 classes, 429 instances with 259 features.
\subsubsection{Average Performance on single/multi- classification}
\label{average_performance}
We test our D-SVM and MD-SVM algorithms compared with standard linear SVM and logistic regression (LG) on Arrhythmia dataset. We run 50 rounds on randomly chosen training datasets and test on the rest datasets. To make sure both the training set and testing set will contain all the classes, we randomly choose the training set in different classes. We choose 80\% data as training data from classes with more than 40 instances and 50\% from classes with less than 40 instances. We test both binary classification and multi classification problems, the result is shown in Table \ref{single cls_nonoise}. Our algorithms perform better than both standard linear SVM and logistic regression on both binary classification and multi classification problems. We have higher accuracy with smaller standard deviation.
\begin{table}
\centering
\caption{Classification Accuracy (\%)}
\label{single cls_nonoise}
    \begin{tabular}{   ccccc}
    \hline
      & SVM   & LG & MD-SVM & D-SVM\\ \hline
    Binary & 88.2$\pm$0.9& 80.4$\pm$1.4 & 89.8$\pm$0.7 & 89.9$\pm$0.7\\
    Multi &  54.6$\pm$3.8 & 40.7$\pm$3.8 & 59.6$\pm$2.5 & 59.8$\pm$2.8\\\hline
    \end{tabular}
\end{table}

\subsubsection{Noise Tolerance Performance}
\begin{figure}
\centering
\includegraphics[width=8.5cm]{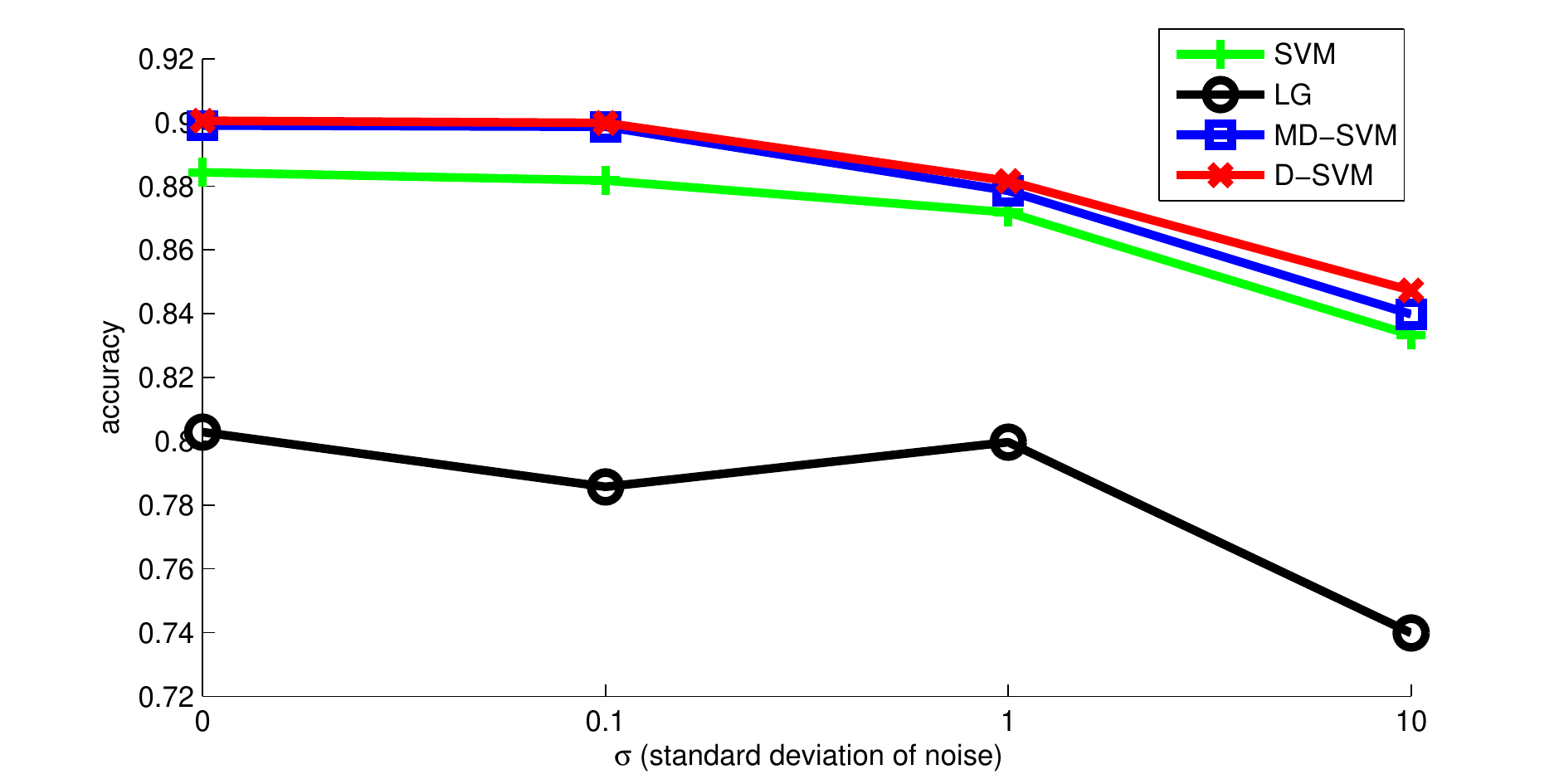}
\caption{Accuracy - $\sigma$ (standard deviation of noise)}
\label{sigmapdf}
\end{figure}

To test the noise tolerance performance of our algorithms, we manually add noise to the data. We have tried different $\sigma$ (standard deviations of noise) as 0, 0.1, 1 and 10. In fact, empirically, the classification performance with $\sigma=10$ is equal to that with $\sigma=\infty$. We randomly choose the training data and test data in the same manner as in Section \ref{average_performance}. The plot of the accuracy against $\sigma$ is shown in Figure \ref{sigmapdf}. Our algorithms always outperform standard linear SVM and logistic regression with $\sigma$ increasing from 0 to 10 (from no artificial noise to infinite amount of artificial noise).
\subsubsection{Effect of Mean Regularization on Dictionary Coefficient $\alpha$}
To test the effect of the mean regularization on dictionary coefficients $\alpha$ introduced in Section \ref{lambda3}, we vary $\lambda_3$ (the weight of mean regularization) from 0 to 1000 as 0, 0.0001, 0.001, 0.01 0.1, 1, 10, 100, 1000 and test on both binary and multi classification tasks. The ranges of accuracy of single and multi classification task are 0.0015 and 0.0014. It indicates that the main contribution to the performance gain is to share the same covariance dictionary $B$ instead of explicitly regularizing the mean of $\alpha_t$.
\subsection{DIGIT RECOGNITION ON MNIST}
\begin{figure}
\centering
\includegraphics[width=8cm]{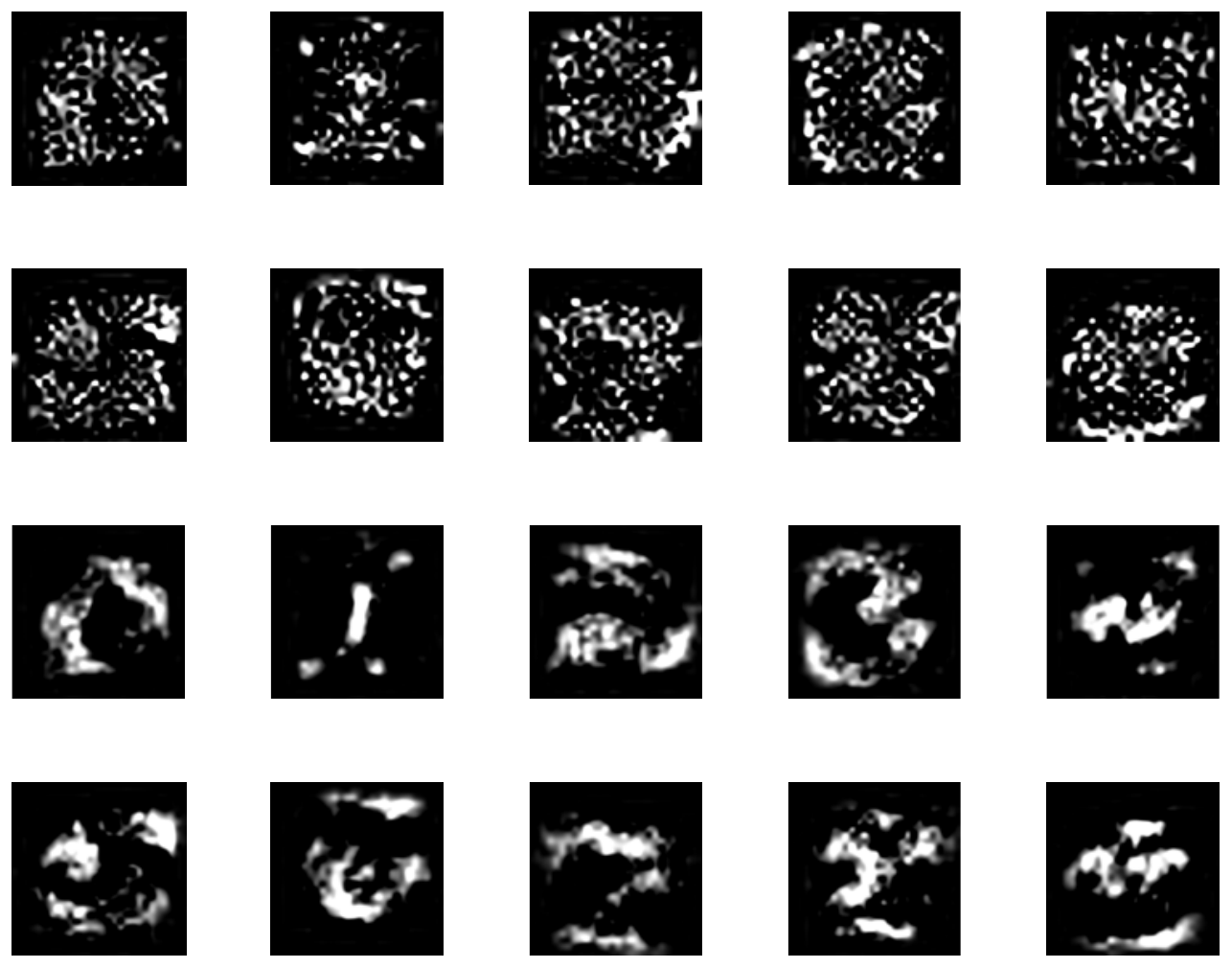}
\caption{Visualization of $w$, the 2 top rows and 2 bottom rows correspond to standard SVM and D-SVM, respectively}
\label{vis_MNIST}
\end{figure}

In this section, we present our experiments on the MNIST handwritten digit recognition benchmark. There are in total 70000 samples in the MNIST dataset of which all samples are normalized 784 dimension vectors vectorized from digit images of size 28$\times$28. Data in MNIST are divided into training and testing sets with 60000 and 10000 samples respectively. \\\\
In this paper, we conduct empirical studies on MNIST dataset with two different settings. The first one is to train class SVMs using both standard linear SVMs and our method (since we have only observed very slight performance difference with and without explicit regularization on the dictionary coefficients $\alpha_t$, we will not show results for the explicitly mean regularized D-SVM in later experiments). The second one is to train instance-based SVMs which means that we have only one positive sample with a large amount of negative samples (also called the ``Exemplar SVM" \cite{malisiewicz-iccv11}).\\\\
In the class SVM setting, we use 10000 training samples to train our models and use all testing samples to test the learned model. In this experiment, the parameter $\lambda_1$ and $\lambda_2$ are set to be 1 and 10, respectively. In the instance-based setting, we use 20000 training samples to calculate a low-dimensional projection basis by PCA to reduce the dimension to 80. Then we use 4000 training samples (400 samples per category) to train 4000 ``Exemplar SVMs", either in the standard way or train D-SVM within each categories. The size of the variance dictionary is set to be 400. The performance of the standard linear SVM and the our method is shown in the table \ref{MNIST_performance}. \\\\
As we can see from the table \ref{MNIST_performance}, even though our method outperforms the standard linear SVM in both settings, the performance gain for the second experimental setting is much more impressive than the first one. There are two main reasons for this phenomenon. The first reason is that we have much more models in the second setting than in the first setting (4000 vs 10), which makes the multi-task knowledge transferring more effective. Another reason is that since we do D-SVM learning for models within a specific category, the models that are learned jointly are naturally highly correlated with each other which is exactly where we should apply multi-task learning. On the other hand, knowledge transferring among non-correlated tasks will actually harm the learning process. Note that this paper is not aiming at the problem of mining the relatedness among different tasks, there are many literatures exploring on topics of finding the correlation among models \cite{kang2011learning,kumar2012learning}.\\\\
SVMs are able to learn discriminative templates from images \cite{malisiewicz-iccv11}, our experiments on MNIST show that our D-SVM could learn image templates with much less noise by leveraging knowledge among different tasks. We can observe from Figure \ref{vis_MNIST} that the templates we learned by D-SVM are with much less noise than templates learned by standard linear SVM.
\begin{table}
\centering
\caption{Classification Error on MNIST (\%)}
\label{MNIST_performance}
    \begin{tabular}{ccc}
    \hline
      & SVM  & D-SVM\\ \hline
    Category & 9.87$\pm$0.21 & 9.07$\pm$0.14\\
    Instance &22.4$\pm$0.64 &16.9$\pm$0.45\\\hline
    \end{tabular}
\end{table}

\subsection{IMAGE CLASSIFICATION ON CALTECH 101}
The Caltech-101 dataset is specific for image classification task \cite{fei2007learning}. The dataset contains 9144 images with 102 categories including vehicles, flowers, animals and background images from google image search. The number of images in different categories varies from 31 to 800. For the accuracy report, we follow the standard experimental protocol for Caltech-101: we select 30 images in each category as training images and test the classification accuracy on the rest of images in each category. For statistical significance, we repeat the experiment for 10 times and calculate the mean accuracy of both standard linear SVM and our D-SVM.\\\\
In order to calculate image descriptors, we first extract the dense SIFT features from images. Then we use k-means algorithm to calculate a visual vocabulary with K entries. Here in this experiment we set K to be 300. Empirical studies suggest typical value of K to be 200-400. Finally we compile the features into a spatial pyramid with level L=3 \cite{lazebnik2006beyond}.\\\\
Having calculated all images into descriptors, we do dimension reduction using PCA to dim=100. Applying both SVM and D-SVM on image descriptors, we have get the classification performance $57.6\pm1.01$ and $58.9\pm0.74$, respectively. Note that we have not made any modification to the features we use but just modify the learning process of linear classifiers to achieve the performance gain.\ref{}

\begin{table}
\centering
\caption{Classification Accuracy on Caltech 101 (\%)}
\label{caltech performance}
    \begin{tabular}{ccc}
    \hline
      & SVM  & D-SVM\\ \hline
     & 57.6$\pm$1.01 & 58.9$\pm$0.74\\\hline
    \end{tabular}
\end{table}
\section{DISCUSSION}
In this paper we have proposed a dictionary based multi-task linear classifier called D-SVM. D-SVM uses a dictionary shared by all tasks to do multi-task knowledge transfer among different tasks. We formally define the learning problem of D-SVM and then present two interpretations of the problem, either from probabilistic and kernel views. From the probabilistic view, we show that the our learning formulation is actually an MAP estimation of all the optimizing variables. It can also be seen as a multiple kernel learning problem since we are trying to find a re-weighting kernel for features from a dictionary of basis (although we are learning a linear classifier). Then, we describe an alternative minimization algorithm to solve this minimization problem and present empirical studies to valid our method.\\\\
In this paper, we have only used the covariance matrix which has only diagonal elements, however, it is possible to use more general covariance matrix to model the learning problem as long as it is positive semi-definite which we will explore in the future.

\bibliographystyle{plain}
\bibliography{mybib}
\end{document}